\pdfoutput=1
\documentclass[letterpaper, 10 pt, conference]{ieeeconf}
\IEEEoverridecommandlockouts
\usepackage{amsmath,amssymb}
\usepackage{graphicx}
\usepackage{booktabs}
\usepackage{cite}
\usepackage{xcolor}
\usepackage{tikz}
\usetikzlibrary{positioning,arrows.meta}
\makeatletter\let\NAT@parse\undefined\makeatother 
\usepackage[hidelinks]{hyperref}
\usepackage{xurl}   
\let\labelindent\relax
\usepackage{enumitem}

\newcommand{\HistMean}{10{,}414}
\newcommand{\HistStd}{1{,}884}
\newcommand{\HistWorst}{5{,}752}
\newcommand{\HistBest}{12{,}327}

\newcommand{\FailGapMax}{5{,}437}
\newcommand{\xgbnoAEFail}{4}
\newcommand{\xgbnoAEN}{15}

\newcommand{\xgbnoAEWorst}{$-$2{,}782}

\newcommand{\xgbAAEN}{15}
\newcommand{\xgbAAEMed}{13{,}393}
\newcommand{\xgbAAEWorst}{12{,}411}

\newcommand{\xgbAAEMv}{0.02}

\newcommand{\xgbAAEOodE}{35}

\newcommand{\xgbAAEwtenkWorst}{13{,}013}
\newcommand{\xgbAAEwtenkTmax}{356.8}

\newcommand{\mtKzeroFail}{13}
\newcommand{\mtKzeroN}{15}
\newcommand{\mtKzeroMed}{$-$261{,}594}

\newcommand{\mtKthreeMed}{13{,}464}
\newcommand{\mtKthreeWorst}{13{,}398}
\newcommand{\mtKthreeTmax}{355.7}

\newcommand{\mtKthreeOod}{0.0}
\newcommand{\mtKthreeOodE}{7.5}

\newcommand{\sgpKzeroAAEwtenkFail}{13}
\newcommand{\sgpKzeroAAEwtenkN}{15}

\newcommand{\KqtrFail}{12}

\newcommand{\KoneMv}{4.31}

\newcommand{\KonehalfMed}{8{,}648}

\newcommand{\KonehalfPoff}{4.7}

\newcommand{\KtwohalfOod}{79}

\newcommand{\KthreefI}{1.32}

\newcommand{\PIfI}{1.77}

\newcommand{\BandLo}{340.1}
\newcommand{\BandHi}{344.4}
\newcommand{\GAAERun}{15}
\newcommand{\GAAEN}{15}
\newcommand{\GAAETmaxLo}{386}
\newcommand{\GAAETmaxHi}{553}

\newcommand{\GmeanPlate}{84}

\newcommand{\XpenMed}{2.9\times10^{-5}}
\newcommand{\XpenMax}{0.01}

\newcommand{\GdivMed}{7}

\newcommand{\LandAAEmax}{0.024}
\newcommand{\LandQrange}{247}
\newcommand{\LandSdRange}{32--1{,}411}

\newcommand{\XnoAEhot}{8}
\newcommand{\XAAEhot}{1}
\newcommand{\XnoAETmed}{362.6}
\newcommand{\XAAETmed}{355.8}

\newcommand{\XrsSevenNo}{13{,}020}
\newcommand{\XrsSevenAAE}{12{,}521}
\newcommand{\XrsZeroAAEtwelve}{13{,}495}
\newcommand{\XrsZeroNotwelve}{4{,}673}

\newcommand{\XfailPlo}{114}
\newcommand{\XfailPhi}{163}
\newcommand{\XfailTlo}{367}
\newcommand{\XfailThi}{375}

\newcommand{\MtKhalfOodE}{42}

\newcommand{\MtKoneOodE}{32}

\newcommand{\MtKtwoOodE}{20}

\newcommand{\MtSweepN}{60}
\newcommand{\MtSweepMedLo}{13{,}464}
\newcommand{\MtSweepMedHi}{13{,}523}

\newcommand{\MtSweepMvMax}{0.04}
\newcommand{\MtSweepTcLo}{341.0}
\newcommand{\MtSweepTcHi}{342.3}
\newcommand{\MtKthreefI}{1.72}

\newcommand{\MtZeroHrs}{10}

\newcommand{\MtSdMed}{4.3}
\newcommand{\MtSigf}{481}

\newcommand{\MtRtwo}{0.978}
\newcommand{\MtCov}{94}
\newcommand{\MtZeroOkN}{2}

\newcommand{\MtZeroOkTlo}{374}
\newcommand{\MtZeroOkThi}{375}
\newcommand{\MtZeroBadTmax}{568}
\newcommand{\MtZeroBadTmin}{553}

\newcommand{\Qhat}{\hat Q}
\newcommand{\bmu}{\mu}
\newcommand{\bsd}{\sigma}
\newcommand{\E}{\mathbb{E}}

\title{\LARGE \bf
Uncertainty-Aware Hybrid Offline Reinforcement Learning with\\
Online Action Selection Enabled by Sparse Gaussian-Process Critics
}

\author{Marco Chan*, Christos Evripidou*, Liqiu Dong,
Benoit Chachuat, and Mehmet Mercang\"oz%
\thanks{* Marco Chan and Christos Evripidou contributed equally to this work.}%
\thanks{All authors are with the Sargent Centre for Process Systems
Engineering, Department of Chemical Engineering, Imperial College London,
London SW7 2AZ, UK.
Corresponding author: M. Mercang\"oz
({\tt\small m.mercangoz@imperial.ac.uk}).}%
}

\begin{document}
\maketitle
\thispagestyle{empty}
\pagestyle{empty}

\begin{abstract}
Start-ups and grade changes remain largely operator-driven, while developing and maintaining dynamic models for model-based control requires substantial engineering effort. Hybrid Offline Learning and Online Optimisation (HOFLON) uses one-step offline reinforcement learning to select actions from historical transition data. In its original implementation, the online objective combines the predicted return from a tree-ensemble critic with an adversarial-autoencoder (AAE) penalty for unfamiliar state--action pairs. We examine the effect of this penalty and replace the critic and AAE with a sparse variational Gaussian-process (sGP) critic. Actions maximise $\mu-\kappa\sigma$, an uncertainty-adjusted estimate of the return from one action followed by the historical operating practice. On the HOFLON polymerisation-reactor benchmark, removing the AAE from HOFLON led to \xgbnoAEFail{} failed start-ups in \xgbnoAEN{} runs over five critics. At its published weight, the variation in the AAE penalty is four orders of magnitude smaller than the variation in the tree-ensemble critic. Nevertheless, it changes actions when the predicted returns of competing candidates are nearly equal, and paired runs with and without the penalty diverged from the first move. With a Matérn-5/2 sGP critic, $\kappa=0$ led to failure in \mtKzeroFail{} of \mtKzeroN{} runs, whereas with $\kappa=3$ none of 15 runs (five critics, three noise realisations) failed. The median and lowest returns were \mtKthreeMed{} and \mtKthreeWorst{} (HOFLON: \xgbAAEMed{} and \xgbAAEWorst). A squared-exponential sGP critic with learned inducing inputs needed $\kappa\ge3$ to remain within the empirical data support in steady state, and then settled with a temperature offset of 2.0\,K.
\end{abstract}
\section{Introduction}

Start-ups, shutdowns and grade changes in process plants remain largely human operator-driven, even where routine operation is automated~\cite{borghesan2022unmanned,baldea2025autonomous}. These transitions can produce off-specification material and take the plant through operating regions where disturbances are more difficult to manage~\cite{durkin2026hoflon}. Process historians contain records of repeated transitions carried out by different operators or under different controller tunings.

Nonlinear model predictive control can automate these transitions, but requires a dynamic model that remains valid throughout the operating range. Developing and maintaining such a model as equipment, feedstocks and operating conditions change involves substantial plant-specific work~\cite{ahmed2025comparative}. Model-free offline reinforcement learning (RL) instead learns from recorded transitions without requiring a process model or exploratory actions on the plant~\cite{levine2020offline}.

Distribution shift remains a central difficulty in offline RL. Maximising a value function fitted to historical data can favour state--action pairs that are poorly represented in the training set~\cite{kumar2019bear,fujimoto2019bcq}. Lee, Kaisare and Lee~\cite{lee2006adp} addressed this problem in approximate dynamic programming with a data-density penalty. Other approaches penalise value uncertainty~\cite{jin2021pessimism,yu2020mopo,an2021edac,ghasemipour2022msg} or correct candidate actions before application~\cite{durkin2025icac}. One-step methods estimate the value of the historical behaviour and perform a policy-improvement step without bootstrapping on unobserved actions~\cite{brandfonbrener2021onestep}.

HOFLON (Hybrid Offline Learning and Online Optimisation)~\cite{durkin2026hoflon} follows this one-step approach (Fig.~\ref{fig:schematic}, top). In its original implementation, a critic is fitted to discounted returns from historical transitions, while an adversarial autoencoder (AAE)~\cite{makhzani2015aae} is trained on logged state--action pairs. At each sampling instant, an optimisation problem selects the action with the highest predicted return, subject to penalties on AAE reconstruction error and the change in the manipulated variables. On a polymerisation-reactor start-up and a paper-machine grade change, HOFLON achieved higher returns than implicit Q-learning~\cite{kostrikov2022iql} and the best historical runs~\cite{durkin2026hoflon}. The AAE requires separate training and introduces a penalty weight whose effective scale depends on the critic and the data.

We evaluate the effect of the AAE on HOFLON's closed-loop behaviour and test a sparse Gaussian-process (GP) critic as an alternative. GP critics provide a predictive mean and standard deviation~\cite{engel2005gptd,savage2021gpq}; the latter can be used to penalise uncertain actions without a separate autoencoder (Fig.~\ref{fig:schematic}, bottom). Sparse GPs were also identified as a possible critic in~\cite{durkin2026hoflon}. A related idea is used in real-time optimisation, where Gaussian processes model the plant--model mismatch and their uncertainty enters the optimisation through acquisition functions and trust regions~\cite{delrio2019modifier,delrio2021rto}. In contrast to the optimistic acquisition functions used there to encourage exploration, such as upper confidence bounds, a lower confidence bound is used here to avoid poorly supported actions, within the existing one-step framework~\cite{jin2021pessimism,srinivas2010gpucb}. The contributions are:
\begin{enumerate}[leftmargin=*,itemsep=1pt,topsep=2pt]
\item an ablation of the AAE over five independently trained HOFLON critics, including an analysis of the AAE penalty relative to the critic values;
\item HOFLON-sGP, which replaces the tree-ensemble critic and AAE with a sparse variational GP critic and uses a lower confidence bound for one-step action selection;
\item a comparison of two sGP critics, with a Matérn-5/2 kernel and fixed inducing inputs and with a squared-exponential kernel and learned inducing inputs, over several uncertainty multipliers $\kappa$, evaluated through closed-loop return, temperature, actuator activity and empirical data support.
\end{enumerate}
Section~\ref{sec:hoflon} reviews HOFLON, Section~\ref{sec:sgp} presents the sGP formulation, and Sections~\ref{sec:setup}--\ref{sec:discussion} describe the benchmark, results and discussion. The case study is made available at \url{https://github.com/AISL-at-Imperial-College-London/HOFLON-sGP}, released as Python code for the simulator, accompanied by the offline dataset, the trained critics, the offline training and online control scripts, and the closed-loop results.

\section{HOFLON}\label{sec:hoflon}

\subsection{Offline learning}
At each sampling instant $k$ the controller receives a measured state $s_k$ and applies an action $a_k\in[0,1]^m$, the vector of scaled manipulated variables. The dataset $\mathcal D$ contains $N_r$ historical transitions of $H$ steps, recorded under unknown and possibly different behaviour policies, for example operators or controllers with different tunings. Every step has a reward $r_k$ and is labelled with its discounted return
\begin{equation}
G_k=\sum_{j=0}^{H-1-k}\gamma^j r_{k+j},\qquad \gamma\in(0,1).\label{eq:G}
\end{equation}
The critic $\Qhat$ is a regression model fitted by least squares to $G_k$ from the features $z_k=[s_k,e_k,\iota_k,a_k]$, where $e_k$ is the set-point error and $\iota_k$ its running integral. It estimates the value $Q_\beta(s,a)$ of applying $a$ in state $s$ and following the historical behaviour afterwards. Because the runs use different controller tunings, $\beta$ denotes the aggregate historical behaviour represented by the dataset. In~\cite{durkin2026hoflon}, $\Qhat$ is a gradient-boosted tree ensemble (XGBoost~\cite{chen2016xgboost}).

The critic may extrapolate poorly to state--action pairs that are absent from the training set. HOFLON therefore also trains an AAE on $x=[s,e,a]$. Its encoder $E$ maps $x$ to a low-dimensional latent code, and its decoder $D$ reconstructs $x$ from the code. The reconstruction error
\begin{equation}
\phi_{\rm lat}(s,a)=\tfrac{1}{n_x}\big\|x-\eta^{-1}\!\big(D(E(\eta(x)))\big)\big\|_2^2,\label{eq:phi}
\end{equation}
where $\eta(\cdot)$ denotes feature-wise scaling, is used as a penalty for state--action combinations that are poorly reconstructed by the AAE.

\subsection{Online optimisation}
At instant $k$ the measured state is fixed and the controller applies
\begin{equation}
a_k^\star=\arg\max_{a\in[0,1]^m}\;\Qhat(z_k(a))-\lambda_1\phi_{\rm lat}(s_k,a)-\lambda_2\|a-a_{k-1}\|_2^2 .\label{eq:hoflon}
\end{equation}
The objective combines predicted return, AAE reconstruction error and a penalty on changes in the manipulated variables. Problem~\eqref{eq:hoflon} has $m$ decision variables, and its objective is not smooth because a tree ensemble is piecewise constant. Three derivative-free methods are used: Nelder--Mead (250 iterations), bounded Powell (250 iterations) and DIRECT (40 iterations). The first two start at $a_{k-1}$; the candidate with the highest objective value is applied. The integrators $\iota$ are frozen while a move is clipped at its bounds (anti-windup)~\cite{durkin2026hoflon}.

\subsection{Relation to offline RL and approximate dynamic programming}\label{sec:onestep}
HOFLON is a one-step offline RL method~\cite{brandfonbrener2021onestep}. The critic is a Monte-Carlo estimate of the value of the behaviour policy, obtained without Bellman backups or iterations. Problem~\eqref{eq:hoflon} is a single greedy policy-improvement step with respect to this estimate. It is restricted by the support term and solved at run time, so no actor network is trained. In the terminology of approximate dynamic programming, this is one iteration of approximate policy iteration, or rollout with the historical operating practice as base policy and with Q-factors estimated from logged data~\cite{bertsekas2019rl}. The method of~\cite{lee2006adp} iterates value iteration on a state cost-to-go and uses a process model for the one-step lookahead; the Q-factor formulation of HOFLON requires no model. The term ``online'' refers to the numerical optimisation of the critic at run time. No learning takes place on the plant.

The one-step approach evaluates each candidate action under the assumption that historical behaviour continues afterwards. It therefore does not optimise a sequence of new actions or directly estimate the optimal policy. Repeating this choice at successive states can combine actions from different historical runs and may achieve returns above those observed in the dataset.

\section{HOFLON-sGP}\label{sec:sgp}

\begin{figure}[t]
\centering
\begin{tikzpicture}[font=\scriptsize, >={Stealth[length=1.6mm]},
  box/.style={draw, rounded corners=1.5pt, align=center, inner sep=2pt, line width=0.4pt},
  hdr/.style={font=\scriptsize\itshape, text=black!65}]
\node[hdr] at (0.72,2.32) {data};
\node[hdr] at (3.02,2.32) {offline models};
\node[hdr] at (6.55,2.32) {online optimisation, every step $k$};
\node[box, fill=gray!12, minimum height=3.45cm, text width=1.12cm] (D) at (0.72,0.30) {Historical\\ transitions\\[2pt] $\mathcal D$\\[2pt] returns\\ $G_k$};
\node[font=\scriptsize\bfseries, anchor=south west, inner sep=0pt] at (1.70,1.88) {HOFLON~\cite{durkin2026hoflon}};
\node[box, fill=blue!9, text width=2.2cm, minimum height=0.72cm] (Q) at (3.02,1.42) {XGBoost critic\\ $\hat Q(s,e,\iota,a)$};
\node[box, fill=blue!9, text width=2.2cm, minimum height=0.72cm] (A) at (3.02,0.52) {AAE\\ $\phi_{\rm lat}(s,e,a)$};
\node[box, fill=blue!4, text width=3.45cm, minimum height=1.62cm] (O1) at (6.55,0.97) {$\max_{a}\ \hat Q-\lambda_1\phi_{\rm lat}-\lambda_2\|\Delta a\|^2$\\[3pt] two learned models;\\ $\lambda_1$ has no fixed unit};
\draw[densely dashed, black!45] (1.62,-0.02) -- (8.35,-0.02);
\node[font=\scriptsize\bfseries, anchor=north west, inner sep=0pt] at (1.70,-0.14) {HOFLON-sGP (this work)};
\node[box, fill=green!12, text width=2.2cm, minimum height=0.72cm] (G) at (3.02,-0.90) {sparse GP critic\\ $\mu,\ \sigma$ on $(s,e,a)$};
\node[box, fill=green!5, text width=3.45cm, minimum height=0.95cm] (O2) at (6.55,-0.90) {$\max_{a}\ \mu-\kappa\sigma-\lambda_2\|\Delta a\|^2$\\[3pt] one model; $\kappa$ = confidence level};
\draw[->] (D.east |- Q) -- (Q.west); \draw[->] (D.east |- A) -- (A.west); \draw[->] (D.east |- G) -- (G.west);
\draw[->] (Q.east) -- (O1.west |- Q.east); \draw[->] (A.east) -- (O1.west |- A.east); \draw[->] (G.east) -- (O2.west);
\end{tikzpicture}
\caption{Structure of HOFLON~\cite{durkin2026hoflon} (top), with a tree-ensemble critic and an AAE, and of HOFLON-sGP (bottom), with one sparse GP critic. In HOFLON-sGP the predictive standard deviation of the critic is the support term.}
\label{fig:schematic}
\end{figure}
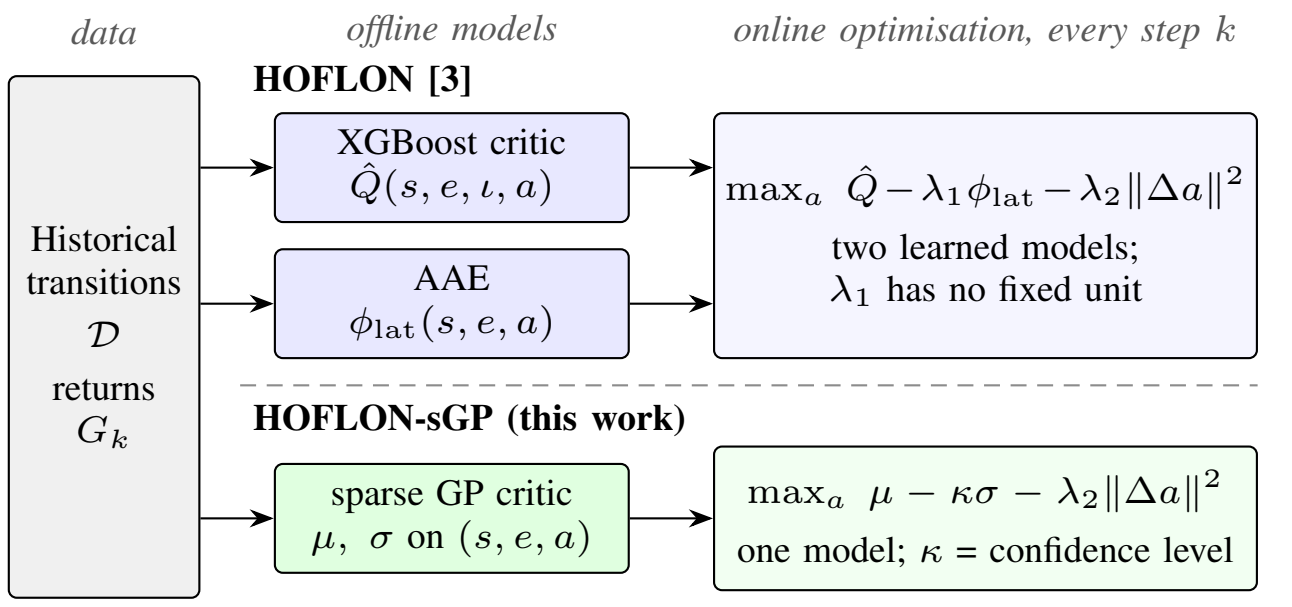

HOFLON-sGP retains the online optimisation in HOFLON but uses a single GP critic to estimate both expected return and predictive uncertainty (Fig.~\ref{fig:schematic}).

\subsection{Sparse variational GP critic}
We place a GP prior on the return function $f(x)$, $f\sim\mathcal{GP}(m,k_\theta)$, which is conditioned on the data. We use a constant prior mean $m$ equal to the mean training return and an automatic-relevance-determination Matérn-5/2 kernel $k_\theta(x,x')=\sigma_f^2\big(1+\sqrt5\,\rho+\tfrac53\rho^2\big)e^{-\sqrt5\rho}$ with $\rho^2=\sum_i (x_i-x_i')^2/\ell_i^2$ on standardised inputs. The squared-exponential kernel $\sigma_f^2e^{-\rho^2/2}$ is used for comparison (Section~\ref{sec:kernel}). The length scales $\ell_i$ determine how fast the return may change along each input. Exact GP regression involves an $N\times N$ matrix factorisation, which is costly for the approximately $2\times10^4$ records used here. We therefore use $M\ll N$ inducing inputs $Z$ to obtain a sparse approximation, as in recent computationally efficient GP models of chemical processes~\cite{dong2026gp}. We use the variational free-energy (VFE) approximation of Titsias~\cite{titsias2009vfe}. The inducing inputs are the $k$-means centres of the standardised training inputs and are kept fixed. Fixed inducing inputs keep the region where $\bsd$ is small tied to the data; in a preliminary study, learning them did not improve the closed-loop behaviour. The hyperparameters $\theta$ and the noise variance $\sigma_n^2$ maximise the collapsed bound
\begin{equation}
\log\mathcal N\!\big(y\,|\,m\mathbf 1,\,Q_{nn}+\sigma_n^2 I\big)-\tfrac{1}{2\sigma_n^2}\operatorname{tr}\!\big(K_{nn}-Q_{nn}\big),
\end{equation}
with $Q_{nn}=K_{nu}K_{uu}^{-1}K_{un}$. With $\Sigma=(K_{uu}+\sigma_n^{-2}K_{un}K_{nu})^{-1}$ and $k_u(x)=k_\theta(Z,x)$, the predictive mean and the latent (epistemic) standard deviation of the return are
\begin{align}
\bmu(x)&=m+\sigma_n^{-2}\,k_u(x)^{\!\top}\Sigma K_{un}(y-m\mathbf 1),\label{eq:mu}\\
\bsd^2(x)&=k_\theta(x,x)-k_u(x)^{\!\top}\big(K_{uu}^{-1}-\Sigma\big)k_u(x).\label{eq:sd}
\end{align}
The latent standard deviation excludes the observation noise $\sigma_n$ and describes uncertainty in the expected return rather than variability between individual returns. It is generally smaller near well-represented inputs. Far from the inducing inputs, $\bsd$ approaches the prior standard deviation $\sigma_f$ and $\bmu$ approaches $m$; the prior mean therefore influences action selection outside the training region. The calibration of $\bsd$ can be checked on held-out logged transitions. Away from the data, $\bsd$ follows from the kernel and its hyperparameters and is a model-based approximation of the epistemic uncertainty. After precomputation, the mean and variance require $\mathcal O(M)$ and $\mathcal O(M^2)$ operations per query, respectively. Their evaluation does not require automatic differentiation at run time.

\subsection{Pessimistic action selection}
HOFLON-sGP replaces~\eqref{eq:hoflon} by
\begin{equation}
a_k^\star=\arg\max_{a\in[0,1]^m}\;\bmu(x_k(a))-\kappa\,\bsd(x_k(a))-\lambda_2\|a-a_{k-1}\|_2^2 ,\label{eq:sgp}
\end{equation}
using the same move penalty and solvers, but no AAE. The critic estimates $Q_\beta$, so the first two terms form a lower confidence bound on the expected return of applying $a$ and then following the historical practice. For a fixed action and a calibrated GP posterior, the posterior probability that the expected return exceeds $\bmu-\kappa\bsd$ is $\Phi(\kappa)$, or 99.87\,\% for $\kappa=3$. This is a pointwise statement. The online solvers evaluate several hundred actions and select the maximiser of $\bmu-\kappa\bsd$, so the pointwise probability does not provide a simultaneous confidence level for the candidates or a guarantee along the closed-loop trajectory. The objective is the pessimistic counterpart of the upper confidence bound used for exploration in GP bandits~\cite{srinivas2010gpucb}. Like support-constrained policy-improvement methods~\cite{laroche2019spibb}, it discourages actions with weak data support, but it does not impose a hard support constraint.

\begin{figure}[t]
\centering
\includegraphics[width=\columnwidth]{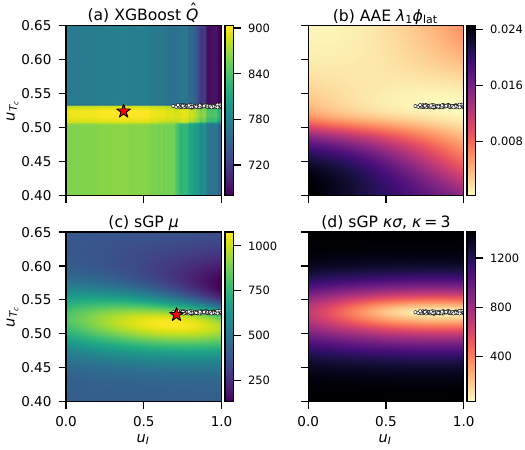}
\caption{Terms of the online objectives over the action set at a logged start-up state (historical run with median return, $t=6$\,h, $P=65$\,kg\,m$^{-3}$). Dots: actions of all 100 historical runs at this time. Stars: maximisers of~\eqref{eq:hoflon} in (a) and of~\eqref{eq:sgp} with $\kappa=3$ in (c). The colour scales differ between panels: the AAE term (b) spans \LandAAEmax{} return units, the critic (a) \LandQrange{} and the sGP term (d) \LandSdRange.}
\label{fig:landscape}
\end{figure}

The penalties in~\eqref{eq:hoflon} and~\eqref{eq:sgp} have different scales and interpretations. The GP term $\kappa\bsd$ is expressed in return units, with a dimensionless multiplier $\kappa$. The AAE term requires a weight $\lambda_1$ to convert reconstruction error into the scale of the objective. The GP standard deviation is determined by the kernel and the training inputs, whereas the AAE penalty depends on reconstruction of each state--action pair. Neither is a direct measure of data density. Fig.~\ref{fig:landscape} shows the terms at a logged start-up state. Near the logged actions both support terms are small. Over the action set, the AAE term varies by \LandAAEmax{} return units, the critic by \LandQrange{} and $\kappa\bsd$ by \LandSdRange.

\subsection{Relation to the historical operating practice}
The following remark relates the performance of~\eqref{eq:sgp} to that of the historical practice. Let $V_\beta(s)=\E_{a\sim\beta(\cdot|s)}Q_\beta(s,a)$ and $A_\beta(s,a)=Q_\beta(s,a)-V_\beta(s)$.

\emph{Remark 1:} Let $\pi$ be the policy defined by~\eqref{eq:sgp}, with $a_{k-1}$ included in the state $s$, and let $\delta(s,a)=\lambda_2\|a-a_{k-1}\|_2^2$. Assume that $|Q_\beta(s,a)-\bmu(s,a)|\le\kappa\bsd(s,a)$ holds at the states visited by $\pi$, for $a=\pi(s)$ and for the actions of $\beta$. Since $\delta\ge0$ and $\pi(s)$ maximises $\bmu-\kappa\bsd-\delta$,
\begin{align*}
Q_\beta(s,\pi(s))&\ge\bmu(s,\pi(s))-\kappa\bsd(s,\pi(s))-\delta(s,\pi(s))\\
&\ge\bmu(s,a)-\kappa\bsd(s,a)-\delta(s,a)\\
&\ge Q_\beta(s,a)-2\kappa\bsd(s,a)-\delta(s,a)
\end{align*}
for every action $a$. Averaging over $a\sim\beta(\cdot|s)$ gives $A_\beta(s,\pi(s))\ge-\E_{a\sim\beta(\cdot|s)}[2\kappa\bsd(s,a)+\delta(s,a)]$. With the performance difference lemma~\cite[Lemma 6.1]{kakade2002approx},
\begin{equation}
V_\pi(s_0)\ge V_\beta(s_0)-\frac{\E_{s\sim d_\pi}\E_{a\sim\beta(\cdot|s)}\big[2\kappa\bsd+\delta\big](s,a)}{1-\gamma},\label{eq:bound}
\end{equation}
where $d_\pi$ is the normalised discounted state distribution of $\pi$ from $s_0$. The same bound holds with $s_0$ replaced by any state of the trajectory.

Under the assumptions of Remark~1, bound~\eqref{eq:bound} relates the loss relative to historical operation to predictive uncertainty and the move penalty at historical actions, evaluated at states visited by $\pi$. The $\bsd$ part is small if these states lie in the support of the data and large otherwise. The share of off-support steps reported in Section~\ref{sec:results} is an empirical indicator of this condition. The remark is a heuristic justification of~\eqref{eq:sgp} and not a performance guarantee for the implemented controller. It assumes that the bounds hold at all visited states, which is most restrictive at $a=\pi(s)$ because $\pi(s)$ maximises $\bmu-\kappa\bsd$, that $\beta$ is a stationary Markov policy in $s$, and that the returns are not truncated. The historical PI controllers depend on their tuning and integrator state, and the returns in~\eqref{eq:G} end with each run.

\subsection{Critic features}
The critic of~\cite{durkin2026hoflon} also receives the integrated errors $\iota_k$. In the historical data the PI controllers saturate during the start-up and their integrators remain frozen. Under the learned controller the integrators continue to integrate, so large integrator values early in a run are outside the data support. The resulting integrated-error features differ between the logged runs and the learned-controller trajectories. A tree ensemble extrapolates such queries with a constant value, and a GP returns its prior. HOFLON-sGP therefore uses the input of the AAE, $x=[s,e,a]$, without integrators. The integrators are part of the state of the historical PI controllers, so this choice also removes information about the continuation policy from the critic. The comparison between HOFLON and HOFLON-sGP changes the critic model, the support term and the critic features together.

\section{Case study and protocol}\label{sec:setup}

\begin{figure*}[t]
\centering
\includegraphics[width=0.95\textwidth]{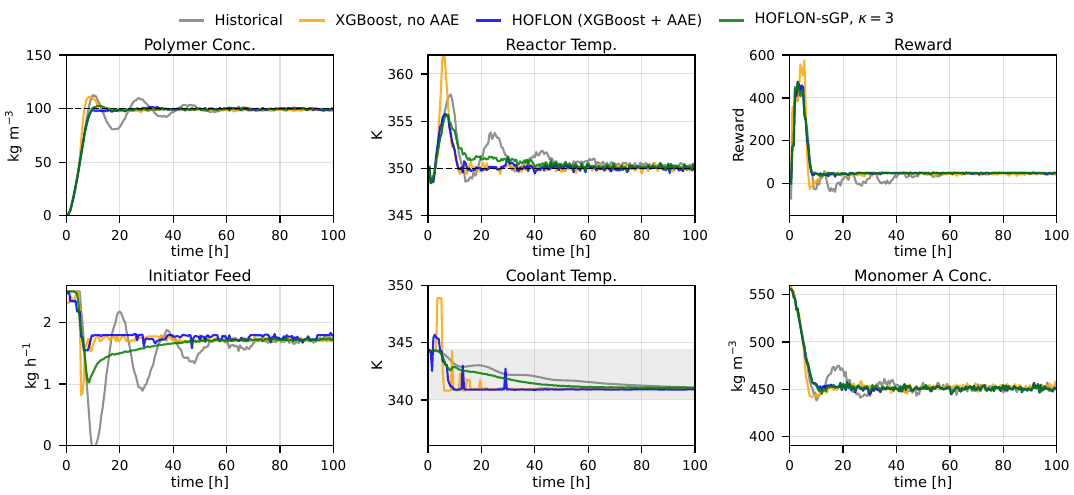}
\caption{One representative run per controller, selected as in~\cite{durkin2026hoflon}. For each learned controller, the run shown has the median return of its 15 runs (five critics $\times$ three noise realisations, seeds 11--13; Section~\ref{sec:eval}). For Historical, the run shown has the median return of the 100 logged runs. Dashed: set-points. Shaded band: range of all logged coolant temperatures.}
\label{fig:median}
\end{figure*}

\subsection{Polymerisation-reactor start-up}
We use the benchmark of~\cite{durkin2026hoflon}: a continuous stirred-tank reactor for free-radical polymerisation based on~\cite{maner1997poly}, with lumped initiation, propagation and termination kinetics and an energy balance with jacket cooling. The reactor starts at 350\,K without initiator, radicals or polymer. The controller has to bring the polymer concentration $P$ to $100$\,kg\,m$^{-3}$ and hold the temperature $T$ at $350$\,K. It manipulates the initiator feed $u_I$ (scaled from $0$--$2.5$\,kg\,h$^{-1}$) and the coolant temperature $u_{T_c}$ (scaled from $280$--$400$\,K). The sampling time is $30$\,min and each run lasts $100$\,h, i.e.\ 200 steps. Monomer, polymer and temperature measurements have Gaussian noise with standard deviations of $2$\,kg\,m$^{-3}$, $0.5$\,kg\,m$^{-3}$ and $0.2$\,K. With $\varepsilon_k=e_{P,k}^2+e_{T,k}^2$ computed from the scaled errors ($P/100$ and $T/350$), the reward is
\begin{equation}
r_k=50\big(e^{-\varepsilon_k/0.05^2}-\varepsilon_k\big)+5000\,(\varepsilon_{k-1}-\varepsilon_k).
\end{equation}
The first term rewards operation near the set-points; the second rewards progress towards them. The reward contains no actuation cost, so actuator activity is assessed separately. The dataset consists of $100$ start-ups under PI control with randomly drawn gains~\cite{durkin2026hoflon}. The total reward is \HistMean$\,\pm\,$\HistStd{} (mean $\pm$ standard deviation; lowest \HistWorst), and the coolant temperature remains within \BandLo--\BandHi\,K in all of them.

\subsection{Critics and weights}
\begin{itemize}[leftmargin=*,itemsep=1pt,topsep=2pt]
\item \emph{XGBoost critics} use the settings of~\cite{durkin2026hoflon}: 1600 trees of depth 8, learning rate 0.03, subsampling 0.9, $\gamma=0.9$, 11 features including the integrators, and a random 80/20 row split, which places rows of the same run in both sets. Five critics differ only in the boosting seed. Their validation $R^2$ is 0.9926--0.9927.
\item \emph{sGP critics} of HOFLON-sGP use the Matérn-5/2 kernel, $M=500$ fixed inducing inputs and $\gamma=0.9$. They are trained on runs 0--94 by maximising the collapsed bound with Adam (learning rate 0.02, 500 iterations, final iterate) and evaluated in the closed form~\eqref{eq:mu}--\eqref{eq:sd}. Five fits differ only in the $k$-means seed. On the held-out runs 95--99 their $R^2$ is \MtRtwo, their 95\,\% predictive intervals contain \MtCov\,\% of the held-out returns, the median latent $\bsd$ is \MtSdMed, and $\sigma_f\approx\MtSigf$.
\item \emph{Squared-exponential sGP critics} are trained on the same runs with GPyTorch~\cite{gardner2018gpytorch}, with the inducing inputs initialised by $k$-means and learned together with the hyperparameters (200 Adam iterations). Five fits differ only in the training seed. On the held-out runs all five have $R^2=0.979$, 94\,\% coverage of the 95\,\% intervals, a median latent $\bsd$ of $3.6$ and $\sigma_f\approx380$. The GP and XGBoost $R^2$ values use different validation splits and should not be compared directly.
\item The \emph{AAE} is the trained network of~\cite{durkin2026hoflon}, with an encoder of layers $9$--$64$--$16$--$4$ with tanh activations and a mirrored decoder.
\item The \emph{weights} of the critic, AAE and move terms in the repository of~\cite{durkin2026hoflon} are $6$, $0.5$ and $0.025$, i.e.\ $\lambda_1=0.083$ and $\lambda_2=0.0042$ in~\eqref{eq:hoflon}. The value of $\lambda_2$ is used for all controllers.
\end{itemize}

\subsection{Controllers compared}
\begin{itemize}[leftmargin=*,itemsep=1pt,topsep=2pt]
\item \emph{HOFLON}: XGBoost critic with the AAE at the published weight, objective~\eqref{eq:hoflon}.
\item \emph{AAE ablations}: the same XGBoost critics without the AAE ($\lambda_1=0$) and with $\lambda_1$ multiplied by $10^4$.
\item \emph{HOFLON-sGP}: Matérn-5/2 sGP critic, objective~\eqref{eq:sgp}, with $\kappa\in\{0,0.5,1,2,3\}$; $\kappa=3$ is the proposed setting.
\item \emph{Squared-exponential sGP critic}: objective~\eqref{eq:sgp} with $\kappa\in\{0,0.25,0.5,1,1.5,2,2.5,3,4,5\}$, and $\kappa=0$ with the AAE term of~\eqref{eq:hoflon} at the published weight and with the weight multiplied by $10^4$.
\end{itemize}

\subsection{Evaluation}\label{sec:eval}
All controllers are evaluated in closed loop on the simulator, with the same initial state and plant parameters. The only random element is the measurement noise. A seed of the random-number generator fixes one realisation of this noise over the 200 steps, so runs with the same seed receive the same noise sequence. Because the controller acts on noisy measurements, different realisations can lead to different actions and trajectories.

A \emph{run} is one controller with one of its five critics on one noise realisation. Differences across critics reflect sensitivity to model fitting, and differences across seeds reflect sensitivity to measurement noise; neither reflects process-model uncertainty. All runs with the same seed are paired, i.e.\ every controller receives the same noise. Each controller is run with each of its five critics on the same three noise realisations (seeds 11--13), which gives 5 critics $\times$ 3 seeds $=$ 15 runs per controller. All results in Section~\ref{sec:results}, including Table~\ref{tab:main} and Figs.~\ref{fig:median}--\ref{fig:kappa}, use these runs. The confidence level $\kappa=3$ was selected on these seeds with the squared-exponential critics, as the smallest tested value for which almost all steps after 20\,h were within the data support. The Matérn-5/2 kernel, the fixed inducing inputs and the number of training iterations were taken from a preliminary study with a separate implementation and other noise realisations.

The following quantities are reported for each run:
\begin{itemize}[leftmargin=*,itemsep=1pt,topsep=2pt]
\item the cumulative reward $R$ of a run. A run is counted as failed if $R<6000$. No historical run has $R$ below \HistWorst, and no failed run in this study has $R$ above \FailGapMax. A low return can result from an offset, an overshoot or a thermal runaway, which are distinguished by the peak temperature and the steady-state values;
\item the peak reactor temperature;
\item the coolant move, defined as the median $|\Delta T_c|$ per step after 20\,h, as a measure of actuator activity in steady state;
\item the steady-state offsets $|\bar P-100|$ and $|\bar T-350|$ of the mean values after 50\,h;
\item the off-support share, defined as the fraction of steps during the start-up (0--20\,h) and after 20\,h at which $x_k$ is outside the data support. A step counts as off-support if the mean standardised distance of $x_k$ to its ten nearest training records exceeds the 99th percentile of the same distance on held-out runs. This is an empirical indicator of support;
\item the solve time of the online optimisation per step.
\end{itemize}

\section{Results}\label{sec:results}

Table~\ref{tab:main} summarises all controllers, Fig.~\ref{fig:box} shows the distributions over runs, and Fig.~\ref{fig:median} shows the run with the median return of each controller.

\subsection{AAE in HOFLON}
With the published weight, HOFLON completed all \xgbAAEN{} runs (median \xgbAAEMed, worst \xgbAAEWorst). Without the AAE, \xgbnoAEFail{} of \xgbnoAEN{} runs failed and the worst return was \xgbnoAEWorst. One critic failed at every seed. In the failed runs the polymer concentration overshot its set-point, with mean values of \XfailPlo--\XfailPhi\,kg\,m$^{-3}$ after 50\,h and peak temperatures of \XfailTlo--\XfailThi\,K. With another critic, the run with seed 12 failed without the AAE (\XrsZeroNotwelve) and reached \XrsZeroAAEtwelve{} with it. The median peak temperature was \XnoAETmed\,K without the AAE and \XAAETmed\,K with it, and \XnoAEhot{} and \XAAEhot{} of the \xgbnoAEN{} runs, respectively, exceeded 360\,K. With one critic, the AAE lowered the median return from \XrsSevenNo{} to \XrsSevenAAE.

At the published weight, the AAE term varies by at most \LandAAEmax{} return units over the action set and the critic by \LandQrange{} (Fig.~\ref{fig:landscape}). At the applied actions, the AAE term had a median of $\XpenMed$ and a maximum of $\XpenMax$. At this scale, the AAE can influence action selection mainly when candidate actions have nearly equal critic values. A tree ensemble has piecewise-constant predictions, so the reconstruction error can distinguish candidate actions within a flat region of the objective. Paired runs with and without the AAE differed from the first move onwards. With the AAE weight multiplied by $10^4$, the AAE term spans a range similar to that of the critic. No run failed, the worst return was \xgbAAEwtenkWorst, and the peak temperature remained below \xgbAAEwtenkTmax\,K.

\subsection{HOFLON-sGP}
\emph{$\kappa=0$.} With the GP mean alone, \mtKzeroFail{} of \mtKzeroN{} runs failed (median return \mtKzeroMed). In these runs the controller set the coolant temperature to its upper bound of 400\,K after a median of \MtZeroHrs\,h, where $\bmu$ equals the prior mean and $\bsd$ the prior standard deviation, and the reactor ran away thermally (peak temperatures of \MtZeroBadTmin--\MtZeroBadTmax\,K). The other \MtZeroOkN{} runs, with one critic, did not fail and reached peak temperatures of \MtZeroOkTlo--\MtZeroOkThi\,K.

\begin{table*}[t]
\caption{Closed-loop results for five critics and three noise realisations (seeds 11--13; 15 paired runs per configuration). A run fails if $R<6000$. Coolant move: median $|\Delta T_c|$ per step after 20\,h. Offsets: median $|\bar P-100|$ and $|\bar T-350|$ after 50\,h. Worst: lowest return over the runs. Off-support: median share of steps outside the data support during the first 20\,h and after 20\,h. The historical row describes the training data.}
\label{tab:main}
\centering\footnotesize\setlength{\tabcolsep}{3.6pt}
\begin{tabular}{@{}c l c r r c c c c c c@{}}
\toprule
Fig.~\ref{fig:box} & Support term in the online objective & Failed & \multicolumn{1}{c}{Median} & \multicolumn{1}{c}{Worst} & Max.\ peak & Coolant & \multicolumn{2}{c}{Steady-state offset} & \multicolumn{2}{c}{Off-support [\%]} \\ \cmidrule(lr){8-9} \cmidrule(l){10-11}
box & & runs & \multicolumn{1}{c}{$R$} & \multicolumn{1}{c}{$R$} & $T$ [K] & move [K] & $P$ [kg\,m$^{-3}$] & $T$ [K] & 0--20\,h & $>$20\,h \\ \midrule
 & \multicolumn{10}{@{}l}{\emph{HOFLON~\cite{durkin2026hoflon}: XGBoost critic $\hat Q$, objective~\eqref{eq:hoflon}}}\\
2 & none & 4/15 & 13{,}020 & $-$2{,}782 & 374.7 & 0.02 & 0.79 & 0.05 & 45 & 0.0\\
3 & AAE, $\lambda_1=0.083$ (published, HOFLON) & 0/15 & 13{,}393 & 12{,}411 & 363.6 & 0.02 & 0.58 & 0.04 & 35 & 0.0\\
4 & AAE, $\lambda_1\times10^4$ & 0/15 & 13{,}372 & 13{,}013 & 356.8 & 0.01 & 0.58 & 0.04 & 22 & 0.0\\
\midrule & \multicolumn{10}{@{}l}{\emph{HOFLON-sGP: sparse-GP critic with Mat\'ern-5/2 kernel, objective~\eqref{eq:sgp}}}\\
 & $\kappa=0$ (GP mean only) & 13/15 & $-$261{,}594 & $-$261{,}752 & 567.7 & 0.00 & 435.03 & 83.58 & 80 & 100\\
5 & $\kappa=3$ (proposed) & \textbf{0}/15 & \textbf{13{,}464} & \textbf{13{,}398} & 355.7 & 0.02 & 0.41 & 0.20 & 7.5 & 0.0\\
\midrule & \multicolumn{10}{@{}l}{\emph{Sparse-GP critic with squared-exponential kernel, objective~\eqref{eq:sgp} with optional AAE term}}\\
 & $\kappa=0$ (GP mean only) & 15/15 & 4{,}267 & 3{,}703 & 359.6 & 5.86 & 16.42 & 1.49 & 80 & 100\\
 & $\kappa=0$ + AAE, $\lambda_1=0.083$ & 15/15 & $-$3{,}351 & $-$38{,}964 & 552.6 & 3.76 & 16.04 & 7.21 & 80 & 100\\
 & $\kappa=0$ + AAE, $\lambda_1\times10^4$ & 13/15 & 4{,}678 & 2{,}932 & 377.9 & 3.07 & 8.03 & 0.74 & 80 & 100\\
6 & $\kappa=3$ & 0/15 & 13{,}378 & 13{,}327 & 355.8 & 0.02 & 0.59 & 1.96 & 48 & 0.6\\
\midrule 1 & Historical PI (100 logged runs) & 4/100 & 11{,}017 & 5{,}752 & 360.3 & 0.01 & 0.29 & 0.08 & 0 & 0\\
\bottomrule
\end{tabular}

\end{table*}

\begin{figure*}[t]
\centering
\vspace{8mm}
\includegraphics[width=1\textwidth]{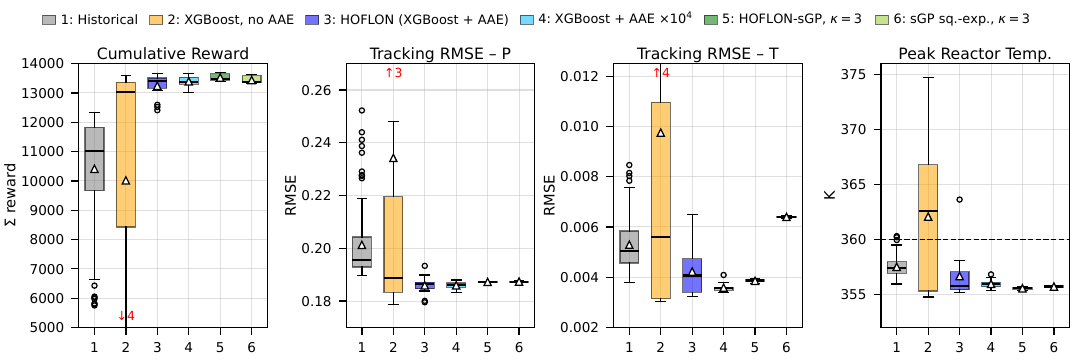}
\caption{Per-run distributions: five critics $\times$ seeds 11--13 per controller (15 runs) and the 100 logged PI start-ups. Boxes span the quartiles, with the median as a black line and the mean as a triangle; whiskers reach 1.5 interquartile ranges. Red arrows count runs outside the plotted range. RMSEs use scaled errors; the dashed line marks 360\,K.}
\label{fig:box}
\end{figure*}

\emph{$\kappa=3$.} On seeds 11--13, the median return was \mtKthreeMed{} (HOFLON: \xgbAAEMed), the lowest return \mtKthreeWorst{} (HOFLON: \xgbAAEWorst), and the peak temperature did not exceed \mtKthreeTmax\,K (Fig.~\ref{fig:box}, boxes 3 and 5). None of the 15 runs failed. During the first 20\,h, \mtKthreeOodE\,\% of the steps were off-support (HOFLON: \xgbAAEOodE\,\%), and after 20\,h \mtKthreeOod\,\%. After 20\,h the median coolant move was 0.02\,K per step (HOFLON: \xgbAAEMv\,K). In steady state, the initiator feed was \MtKthreefI\,kg\,h$^{-1}$ (PI: \PIfI\,kg\,h$^{-1}$), and the offsets were 0.41\,kg\,m$^{-3}$ and 0.20\,K (Fig.~\ref{fig:median}).

\subsection{Confidence level}
\begin{figure}[t]
\centering
\includegraphics[width=\columnwidth]{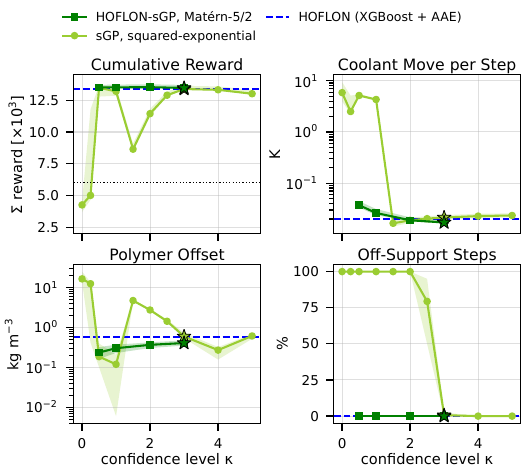}
\caption{Effect of the confidence level $\kappa$ (five critics per kernel, seeds 11--13; line: median, band: range; stars: $\kappa=3$). Matérn-5/2 critics for $\kappa\ge0.5$; $\kappa=0$ in Section~\ref{sec:results}-B. Coolant move: median $|\Delta T_c|$ per step after 20\,h. Polymer offset: $|\bar P-100|$ after 50\,h. Off-support: share of steps after 20\,h. Dashed: HOFLON on the same runs.}
\label{fig:kappa}
\end{figure}

With the Matérn-5/2 critics and $\kappa\in\{0.5,1,2,3\}$, none of the \MtSweepN{} runs on seeds 11--13 failed, and the median returns were \MtSweepMedLo--\MtSweepMedHi{} (Fig.~\ref{fig:kappa}). No step after 20\,h was off-support in any of these runs, the coolant temperature after 20\,h remained within \MtSweepTcLo--\MtSweepTcHi\,K, and the median coolant move was at most \MtSweepMvMax\,K per step. During the first 20\,h, the median share of off-support steps was \MtKhalfOodE\,\% at $\kappa=0.5$, \MtKoneOodE\,\% at $\kappa=1$, \MtKtwoOodE\,\% at $\kappa=2$ and \mtKthreeOodE\,\% at $\kappa=3$.

\subsection{Squared-exponential kernel}\label{sec:kernel}
\emph{$\kappa=0$.} All 15 runs failed (median 4{,}267). Within about 8\,h, the controller reached states where $\bmu$ and $\bsd$ were effectively constant across the action set, close to their prior values. The online objective was then dominated by the move penalty, so the warm-started searches retained the previous move and the polymer concentration settled near \GmeanPlate\,kg\,m$^{-3}$. With the AAE at the published weight, the paired runs separated after a median of \GdivMed{} steps. At the states where the GP is constant, the AAE term is the only term of the objective that depends on the action, and it selects actions by reconstruction error for states outside the data. \GAAERun{} of \GAAEN{} runs ended in thermal runaway, with peak temperatures of \GAAETmaxLo--\GAAETmaxHi\,K. With the AAE weight multiplied by $10^4$, \sgpKzeroAAEwtenkFail{} of \sgpKzeroAAEwtenkN{} runs failed.

\emph{$\kappa>0$.} For $\kappa\le0.25$ most runs failed (\KqtrFail{} of 15 at $\kappa=0.25$). For $0.5\le\kappa\le2$ the median share of off-support steps after 20\,h was 100\,\%, and at $\kappa=2.5$ it was \KtwohalfOod\,\%. For $\kappa\le1$ the coolant temperature changed by several kelvin per step (\KoneMv\,K at $\kappa=1$), and for $1.5\le\kappa\le2.5$ the inputs were smooth and the polymer offset reached \KonehalfPoff\,kg\,m$^{-3}$ (median return \KonehalfMed{} at $\kappa=1.5$). With $\kappa=3$, none of the 15 runs of the squared-exponential controller failed (median 13{,}378, lowest 13{,}327), and all but 0.6\,\% of the steps after 20\,h were within the data support. This controller is not shown in Fig.~\ref{fig:median}. It settled on a branch with an initiator feed of \KthreefI\,kg\,h$^{-1}$ (PI: \PIfI\,kg\,h$^{-1}$) and a reactor temperature 2.0\,K above the set-point (Fig.~\ref{fig:box}, box 6). At steady-state states visited with $\kappa=0.5$, the maximum of $\bmu$ was 200--400 return units above $\bmu$ at the logged actions, and $\bsd$ at the maximiser was about 200 compared with less than 30 at the logged actions. The maximiser of $\bmu-\kappa\bsd$ lay in the logged coolant range only for $\kappa$ between 3 and 5.

\subsection{Computation}
HOFLON-sGP does not require the AAE evaluation during online optimisation, and its solve time per step was shorter than that of HOFLON. All computations were run on one core of an Intel Core Ultra 5 135U processor. For both controllers, the solve time per step was far below the 30\,min sampling interval.

\section{Discussion}\label{sec:discussion}

\emph{Support terms.} The AAE reconstruction error and the GP standard deviation can both penalise poorly represented state--action pairs, although neither is necessarily monotonic in distance to the training set. The GP term is expressed in return units, with $\kappa$ a number of predictive standard deviations; a multiple of the GP standard deviation is used in the same way to tighten constraints in~\cite{hewing2020cautious}. The effective strength of $\lambda_1$ depends on the reconstruction scale, the reward scale and the critic. At the published value, the AAE term matters mainly where the critic is flat, which for a tree ensemble arises from its piecewise-constant predictions, including outside the training domain. Scaling it by $10^4$ removed the failures of the tree critic but not those of the squared-exponential GP mean.

\emph{Actuator activity.} The reward contains no actuation cost. With $\lambda_2=0.0042$, a change of 0.05 in $u_{T_c}$ costs $\lambda_2\cdot0.0025$, which is small compared with differences of hundreds of return units between critic values. With the squared-exponential critic and $\kappa\le1$, the returns were high and the coolant temperature changed by several kelvin per step outside the logged range. In HOFLON with the XGBoost critic, smooth steady-state actuation results from the constant extrapolation of the tree ensemble. With the sGP critics, it was obtained for all tested $\kappa\ge0.5$ with the Matérn-5/2 kernel and for $\kappa\ge1.5$ with the squared-exponential kernel. The two kernels had similar accuracy on the held-out logged runs and different closed-loop behaviour for $\kappa<3$.

\emph{One-step improvement.} Both controllers select actions through a one-step improvement objective relative to historical behaviour (Section~\ref{sec:onestep}). At suitable settings, they can achieve returns above the best recorded historical run (\HistBest), but this does not hold for every critic or penalty setting. Further improvement would require the evaluation of the new policy and additional improvement steps, for example by fitted Q-evaluation with $\bsd$ restricting the updates to the data support. This reintroduces bootstrapping.

\emph{Use in autonomous operation.} HOFLON-sGP uses a single learned critic, which could be retrained when additional transitions become available. Its predictive standard deviation is also available online and could provide a supervisory indication of poorly supported actions~\cite{borghesan2022unmanned,baldea2025autonomous}.

\emph{Limitations.} The study uses one simulator with a fixed initial state, and the seeds vary only the measurement noise. Robustness to plant parameters and disturbances - especially nonstationary disturbances - has not been tested. The XGBoost critics use the integrators and the sGP critics do not, and an XGBoost critic without integrators was not evaluated. Without integrated-error features, the action-selection law has no explicit integral compensation, and steady-state offsets can persist. The confidence level was selected with the squared-exponential critics on the same three noise seeds on which all results are reported. The two kernel implementations also use different inducing-input treatments and training schedules, so their closed-loop differences cannot be attributed to the kernel alone. The paper-machine benchmark of~\cite{durkin2026hoflon}, with four manipulated variables and no measurements of internal states, remains to be evaluated; it would also provide a test of the required number of inducing points.

\section{Conclusion}
HOFLON is a one-step offline RL method, and in this study its tree-ensemble critic and AAE were replaced by one sparse GP critic with a lower-confidence-bound objective. With a Matérn-5/2 critic and $\kappa=3$, all 15 runs over five critics and three noise realisations completed without failure. The median return was comparable to that of HOFLON and the lowest return higher, with smooth actuation and peak temperatures below 356\,K. Without a support term, a large share of runs failed with both the tree-ensemble critic and the GP mean; in HOFLON-sGP, the $\bsd$ term prevents these failures without a separate model. In HOFLON, the AAE term at its published weight changes the selected action mainly where the tree-ensemble critic is nearly constant.

\section*{ACKNOWLEDGMENT}
\noindent\begin{minipage}{\columnwidth}
\setlength{\parindent}{1em}\indent The authors used Claude Opus 5.5 (Anthropic) to draft and revise parts of the text, mainly in the abstract, Section~\ref{sec:eval} and Sections~\ref{sec:results}--\ref{sec:discussion}, and to check the consistency of the reported results. All content was reviewed and edited by the authors, who take full responsibility for the paper.
\end{minipage}


\end{document}